\documentclass[11pt,a4paper]{article}
\PassOptionsToPackage{breaklinks}{hyperref}
\usepackage[hyperref]{acl2019}
\usepackage{times}
\usepackage{latexsym}
\usepackage{amssymb}
\usepackage{amsmath}
\usepackage{booktabs} 
\usepackage{enumitem}
\usepackage{multirow}
\PassOptionsToPackage{breaklinks}{url}
\usepackage{url}
\usepackage{graphicx}
\usepackage{subfigure}
\pdfoutput=1

\aclfinalcopy 
\def\aclpaperid{104} 

\newif\ifdraft
\draftfalse

\DeclareMathOperator{\nn}{NN}

\usepackage{siunitx}

\newcommand{\Ni}{({\em i})~}
\newcommand{\Nii}{({\em ii})~}
\newcommand{\Niii}{({\em iii})~}
\newcommand{\Niv}{({\em iv})~}

\newcommand{\SiEn}{Sinhala--English~}

\newcommand{\NeEn}{Nepali--English~}

\newcommand{\SiEnns}{Sinhala--English}
\newcommand{\NeEnns}{Nepali--English}

\newcommand{\laser}{LASER{}}
\newcommand{\fairseq}{{\ttfamily{fairseq{}}}}
\newcommand{\flores}{{\ttfamily{flores{}}}}

\title{Low-Resource Corpus Filtering using Multilingual Sentence Embeddings}

\author{
{\bf Vishrav Chaudhary}$^\blacklozenge$ \enskip 
{\bf Yuqing Tang}$^\blacklozenge$ \enskip 
{\bf Francisco Guzm\'{a}n$^{\blacklozenge}$}
\enskip {\bf Holger Schwenk}$^{\blacklozenge}$
{\bf Philipp Koehn}$^{\blacksquare}$   \\
$^{\blacklozenge}$Facebook AI \enskip 
$^{\blacksquare}$Johns Hopkins University  \\
\texttt{\{vishrav,yuqtang,fguzman,schwenk\}@fb.com} \enskip \texttt{phi@jhu.edu}\\
}

\date{}

\begin{document}
\maketitle

\begin{abstract}

In this paper, we describe our submission to the WMT19 low-resource parallel corpus filtering shared task. Our main approach is based on the \laser\ toolkit (Language-Agnostic SEntence Representations), which uses an encoder-decoder architecture trained on a parallel corpus to obtain multilingual sentence representations.
We then use the representations directly to score and filter the noisy parallel sentences without additionally training a scoring function. We contrast our approach to other promising methods and show that \laser{} yields strong results. Finally, we produce an ensemble of different scoring methods and obtain additional gains. Our submission achieved the best overall performance for both the \NeEn and \SiEn $1$M tasks by a margin of $1.3$ and $1.4$ BLEU respectively, as compared to the second best systems. Moreover, our experiments show that this technique is promising for low and even no-resource scenarios.

\end{abstract}

\section{Introduction} \label{sec:intro}

The availability of high-quality parallel training data is critical for obtaining good translation performance, as
neural machine translation (NMT) systems are less robust against noisy parallel data than statistical machine translation (SMT) systems \cite{khayrallah2018impact}. Recently, there is an increased interest in the filtering of noisy parallel corpora (such as Paracrawl\footnote{\url{http://www.paracrawl.eu/}}) to increase the amount of data that can be used to train translation systems \cite{koehn2018findings}.  

While the state-of-the-art methods that use NMT models have proven effective in mining parallel sentences \cite{junczys2018dual} for high-resource languages, their effectiveness has not been tested in low-resource languages. The implications of low availability of training data for parallel-scoring methods is not known yet.

For the task of low-resource filtering \cite{filtering:2019:WMT}, we are provided with a very noisy $40.6$ million-word (English token count) \NeEn corpus and a $59.6$ million-word \SiEn corpus crawled from the web as part of the Paracrawl project. The challenge consists of providing scores for each sentence pair in both noisy parallel sets. The scores will be used to subsample sentence pairs that amount to $1$ million and $5$ million English words. The quality of the resulting subsets is determined by the quality of a statistical machine translation (Moses, phrase-based \cite{moses}) and the neural machine translation system \fairseq{} \cite{ott2019fairseq} trained on this data. The quality of the machine translation system will be measured by BLEU score using SacreBLEU \cite{post2018sacrebleu} on a held-out test set of Wikipedia translations for \SiEn and \NeEn from the \flores{} dataset \cite{flores2019}.

In our submission for this shared task, we use of multilingual sentence embeddings obtained from 
{\laser}\footnote{\url{https://github.com/facebookresearch/LASER}} which uses an encoder-decoder architecture to train a multilingual sentence representation model using a relatively small parallel corpus. 
Our experiments demonstrate that the proposed approach outperforms other existing approaches. Moreover we make use of an ensemble of multiple scoring functions to further boost the filtering performance.

\section{Methodology} \label{sec:methodology}

The WMT 2018 shared task for parallel corpus filtering \cite{koehn2018findings}\footnote{\tt http://statmt.org/wmt18/ parallel-corpus-filtering.html} introduced several methods 
to tackle a high-resource German-English data condition.
While many of these methods were successful to filter out noisy translations, few have been tried under low-resource conditions. In this paper, we address the problem of low-resource sentence filtering using sentence-level representations and compare them to other popular methods used in high-resource conditions. 

The \laser{} model \cite{artetxe2018margin} makes use of multilingual sentence representations to gauge the similarity between the source and the target sentence. It has provided  state-of-the-art performance on the BUCC corpus mining task and has also been effective in filtering WMT Paracrawl data \cite{artetxe2018margin}. However, these tasks only considered high-resource languages, namely French, German, Russian and Chinese. Fortunately, this technique has also been effective on zero-shot cross-lingual natural language inference in the XNLI dataset \cite{artetxe2018multi} which makes it promising for the low resource scenario being focused in this shared task. In this paper, we propose to use an adaptation of \laser{} to low-resource conditions to compute the similarity scores to filter out noisy sentences. 
\\
For comparison to \laser{}, we also establish initial benchmarks using Bicleaner and Zipporah, two popular baselines which have been used in the Paracrawl project; and dual conditional cross-entropy, which has proven to be state-of-the-art for the high-resource corpus filtering task \cite{koehn2018findings}. We explore the performance of the techniques under similar pre-processing conditions regarding language identification filtering and lexical overlap. We observe that \laser{} scores provide a clear advantage for this task. Finally, we perform ensembling of the scores coming from different methods. We observe that when \laser{} scores are included in the mix, the boost in performance is relatively minor. In the rest of this section we discuss the settings for each of the methods applied.

\subsection{\laser{} Multilingual Representations}
The underlying idea is to use the distances between two multilingual representations as a notion of parallelism between the two embedded sentences \cite{Schwenk:2018:acl_mine}. To do this, we first train an encoder that learns to produce a multilingual, fixed-size sentence representation; and then compute a distance between two sentences in the learned embedding space. In addition, we use a \emph{margin} criterion, which uses a $k$ nearest neighbors approach to normalize the similarity scores given that  cosine similarity is not globally consistent \cite{artetxe2018margin}.\\

\paragraph{Encoder}  The multilingual encoder consists of a bidirectional LSTM, and our sentence embeddings are obtained by applying max-pooling over its output.  
We use a single encoder and decoder in our system, which are shared by all languages involved. For this purpose, we trained multilingual sentence embeddings on the provided parallel data only (see Section~\ref{SectTrain} for details).
\\

\paragraph{Margin} We follow the definition of  \emph{ratio}\footnote{We explored the \emph{absolute}, \emph{distance} and \emph{ratio} margin criteria, but the latter worked best} from \citep{artetxe2018margin}. Using this, the similarity score between two sentences (x, y) can be computed as 
\begin{equation}
    \frac{ 2k\cos(x, y)}{
    \sum_{y' \in \nn_k(x)}{\cos(x, y')} +  \sum_{x' \in \nn_k(y)}{\cos(x', y)}) }
    \nonumber
\end{equation}
where $\nn_k(x)$ denotes the $k$ nearest neighbors of $x$ in the other language, and analogously for $\nn_k(y)$. Note that this list of nearest neighbors does not include duplicates, so even if a given sentence has multiple occurrences in the corpus, it would have (at most) one entry in the list.


\paragraph{Neighborhood} Additionally, we explored two ways of sampling $k$ nearest neighbors. First a \emph{global} method, in which we used the neighborhood comprised of the noisy data along with the clean data. Second a  \emph{local} method, in which we only scored the noisy data using the noisy neighborhood, or the clean data using the clean neighborhood.\footnote{this last part was only done for training an ensemble} 

\subsection{Other Similarity Methods}

\paragraph{\bf Zipporah}  \cite{xu2017zipporah,khayrallah-xu-koehn:2018:WMT}, which is often used as a baseline comparison, uses language model and word translation scores, with weights optimized to separate clean and synthetic noise data.
In our setup, we trained Zipporah models for both language pairs \SiEn{} and \NeEnns. We used the open source release\footnote{\url{https://github.com/hainan-xv/zipporah}} of the Zipporah tool without modifications. All components of the Zipporah model (probabilistic translation dictionaries and language models) were trained on the provided clean data (excluding the dictionaries). Language models were trained using KenLM \cite{Heafield-estimate} over the clean parallel data. We are not using the provided monolingual data, as per default setting. We used the  development set from the \flores{} dataset for weight training.

\paragraph{\bf Bicleaner} \cite{sanchez2018prompsit} uses lexical translation and language model scores, and several shallow features such as: respective length, matching numbers and punctuation.
As with Zipporah, we used the open source Bicleaner\footnote{\url{https://github.com/bitextor/bicleaner}} toolkit unmodified out-of-the-box. Only the provided clean parallel data was used to train this model. Bicleaner uses a rule-based component to identify noisier examples in the parallel data and trains a classifier to learn how to separate them from the rest of the training data. 
The use of language model features is optional.  
We only used models without a language model scoring component.\footnote{We found that 
including a LM as a feature resulted in almost all sentence pairs receiving a score of 0.}

\paragraph{Dual Conditional Cross-Entropy}
One of the best performing methods on this task was {dual conditional cross-entropy filtering} \cite{junczys2018dual}, which uses a combination of forward and backward models to compute a cross-lingual similarity score. 
In our experiments, for each language pair, we used the provided clean training data to train neural machine translation models in both translation directions: source-to-target and target-to-source. Given such a translation model $M$, we force-decode sentence pairs $(x,y)$ from the noisy parallel corpus and obtain the cross-entropy score
\begin{equation}
H_M(y|x) = \frac{1}{|y|} \sum_{t=1}^{|y|} \log p_M(y_t|y_{[1,t-1]},x)
\end{equation}

Forward and backward cross entropy scores, $H_F(y|x)$ and $H_B(x|y)$ respectively, are then averaged with an additional penalty on a large difference between the two scores $|H_F(y|x) - H_B(x|y)|$.
\begin{align}
 \textrm{score}(x, y) &=  \frac{H_F(y|x) + H_B(x|y)}{2} \\
 &\qquad   -| H_F(y|x) - H_B(x|y)|  \nonumber
\end{align}

The forward and backward models are five-layer encoder/decoder transformers trained using \fairseq{} with parameters identical to the ones used in the baseline \flores{} model \footnote{\tt https://github.com/facebookresearch/flores \#train-a-baseline-transformer-model}. The models were trained on the clean parallel data for $100$ epochs. For the Nepali-English task, we also explored using Hindi-English data without major differences in results. We used the \flores{} development set to pick the model that maximizes BLEU scores.

\subsection{Ensemble}
To leverage over the strengths and weaknesses of different scoring systems, we explored the use of a binary classifier to build an ensemble. While it's trivial to obtain positives (e.g. the clean training data), mining negatives can be a daunting task. Hence, we use positive-unlabeled (PU) learning \cite{ mordelet2014bagging}, which allows us to obtain classifiers without having to curate a dataset of explicit positive and negatives. In this setting our positive labels come from the clean parallel data while the unlabeled data comes from the noisy set.

To achieve this, we apply bagging of  $100$ weak, biased classifiers (i.e. with a 2:1 bias for unlabeled data vs. positive label data). We use support vector machines (SVM) with a radial basis kernel, and we randomly sub-sample the set of features for training each base classifier, helping keep them diverse and low-capacity. 

We ran two iterations of training of this ensemble. In the first iteration we used the original positive and unlabeled data described above. For the second iteration, we used the learned classifier to re-label the training data. We explored several re-labeling approaches (e.g. setting a threshold that maximizes $F_1$ score). However, we found that setting a class boundary to preserve the original positives-to-unlabeled ratio worked best. We also observed that the performance deteriorated after two iterations. 

\section{Experimental Setup} \label{sec:experiments}
We experimented with various methods using a setup that closely mirrors the official scoring of the shared task. 
All methods are trained on the provided clean parallel data (see Table~\ref{tab:resources}). We did not use the given monolingual data. For development purposes, we used the provided \flores{} dev set. 
For evaluation, we trained machine translation systems on the selected subsets ($1$M, $5$M) of the noisy parallel training data using \fairseq{} with the default \flores{} training parameter configuration. We report  SacreBLEU scores on the \flores{} devtest set. We selected our main system based on the best scores on the devtest set for the $1$M condition.

\begin{table}[h]
    \centering \small
    \begin{tabular}{lccc}
        \toprule
        & {\bf si-en} & {\bf ne-en} & {\bf hi-en} \\
        \midrule
        Sentences & 646k & 573k & 1.5M  \\
        English words & 3.7M & 3.7M & 20.7M\\
        \bottomrule
    \end{tabular}
    \caption{Available bitexts to train the filtering approaches.}
    \label{tab:resources}
\end{table}

\vspace{-0.3cm}
\subsection{Preprocessing}
 We applied a set of filtering techniques similar to the ones used in \laser{} \cite{artetxe2018margin} and assigned a score of $-1$ to the noisy sentences based on incorrect language on either the source or the target side or having an overlap of at least 60\% between the source and the target tokens. We used fastText\footnote{\tt https://fasttext.cc/docs/en/language- identification.html} for language id filtering. Since \laser{} computes similarity scores for a sentence pair using these filtering techniques, we experimented by adding these to the other models we used for this shared task.


\subsection{\laser{} Encoder Training}
\label{SectTrain}


For our experiments and the official submission, we trained a multilingual sentence encoder using the permitted resources in  Table~\ref{tab:resources}.
We trained a single encoder using all the parallel data for \SiEnns{}, \NeEnns{} and Hindi-English. Since Hindi and Nepali share the same script, we concatenated their corpora into a single parallel corpus. To account for the difference in size of the parallel training data, we over-sampled the \SiEn and Nepali/Hindi-English bitexts in a ratio of $5$:$3$. This resulted in roughly $3.2$M training sentences for each language direction, i.e. Sinhala and combined Nepali-Hindi.
The models were trained using the same setting as the public \laser{} encoder which involves normalizing  texts and tokenization with Moses tools (falling back to the English mode). We first learn a joint $50$k BPE vocabulary on the concatenated training data using fastBPE\footnote{\url{https://github.com/glample/fastBPE}}. The encoder sees Sinhala, Nepali, Hindi and English sentences at the input
, without having any information about the current language. This input is always translated into English.\footnote{This means that we have to train an English auto-encoder. This didn't seem to hurt, since the same encoder also handles the three other languages}
We experimented with various techniques to add noise to the English input sentences, similar to what is used in unsupervised neural machine translation, e.g. \cite{unsupNMTartetxe,lample_emnlp2018}, but this did not improve the results.

The encoder is a five-layer BLSTM with $512$ dimensional layers. The LSTM decoder has one hidden layer of size $2048$, trained with the Adam optimizer. For development, we calculate similarity error on the concatenation of the \flores{} dev sets for \SiEn and \NeEnns.
Our models were trained for seven epochs 
for about $2.5$ hours on $8$ Nvidia GPUs.

\section{Results}

\begin{table}[hbt]
\centering
\small

\begin{tabular}{p{2.7cm}llll}

\toprule
\bf Method & \multicolumn{2}{c}{\bf  ne-en} & \multicolumn{2}{c}{\bf  si-en} \\\cmidrule(r){2-3}\cmidrule(l){4-5}
& \bf 1M & \bf 5M & \bf 1M & \bf 5M\\
\midrule
{\bf Zipporah}  \\
base & 5.03 & 2.09 & 4.86 & 4.53\\
\quad+ LID & 5.30 & 1.53 & 5.53 & 3.16 \\
\quad\quad+ Overlap & 5.35 & 1.34 & 5.18 & 3.14  \\\midrule
{\bf Dual X-Ent.} \\
base & 2.83 & 1.88 & 0.33 & 4.63$^+$ \\
\quad+ LID & 2.19 & 0.82 & 6.42 & 3.68 \\
\quad\quad+ Overlap &  2.23 & 0.91 & 6.65 & 4.31  \\\midrule
{\bf Bicleaner}\\
base & 5.91 & 2.54$^+$ & 6.20 & 4.25 \\
\quad+ LID & 5.88 & 2.09 & 6.36 & 3.95 \\
\quad\quad+ Overlap & 6.12$^+$ & 2.14 & 6.66$^+$ & 3.26  \\\midrule
{\bf \laser{}} \\
\quad \emph{local} & \emph{7.37*} & {\bf 3.15} & \emph{7.49*} & 5.01 \\ 
\quad \emph{global} & 6.98 & \emph{2.98*} & 7.27 & 4.76 \\\midrule
{\bf Ensemble} \\
ALL & 6.17 & 2.53 & {\bf 7.64} & {\bf 5.12} \\ 
\laser{} \emph{glob. + loc.} & {\bf 7.49} & 2.76 & 7.27 & \emph{5.08*} \\ 
\bottomrule
\end{tabular}
\caption{SacreBLEU scores on the \flores{} devtest set. In {\bf bold}, we highlight the best scores for each condition. In \emph{italics*}, we highlight the runner up. We also signal the best non-\laser{} method with $^+$.}
\label{tab:results}
\end{table}

From the results in Table~\ref{tab:results}, we observe several trends: \Ni the scores for the $5$M condition are generally lower than for the $1$M condition. This condition appears to be exacerbated by the application of language id and overlap filtering. 
\Nii \laser{} shows consistently good performance. The \emph{local} neighborhood works better than the \emph{global} one. In that setting, \laser{} is on average $0.71$ BLEU above the best non-\laser{} system. These gaps are higher for the $1$M condition ($0.94$ BLEU).
\Niii The best ensemble configuration provides small improvements over the best \laser{} configuration. For \SiEn the best configuration includes every other scoring method (ALL). For \NeEn the best configuration is an ensemble of \laser{} scores. 
\Niv  Dual cross entropy shows mixed results. For \SiEnns, it only works once the language id filtering is enabled which is consistent with previous observations \cite{junczys2018dual}. For \NeEnns, it provides scores well below the rest of the scoring methods. 
Note that we did not perform an architecture exploration. 

\paragraph{Submission} For the official submission, we used the \emph{ALL} ensemble for the \SiEn task and the \laser{} \emph{global + local} ensemble for the \NeEn task. We also submitted the \laser{} \emph{local} as a contrastive system. As we can see in Table~\ref{tab:official}, the results from the main and contrastive submissions are very close. In one case, the contrastive solution (a single \laser{}) model yields better results than the ensemble. These results placed our 1M submissions 1.3 and 1.4 BLEU points above the runner ups for the \NeEn and \SiEn tasks, respectively. As noted before, our systems perform worse on the 5M condition. We also noted that the numbers in Table~\ref{tab:results} differ slightly from the ones reported in \cite{filtering:2019:WMT}. We attribute this difference to the effect of training in $4$ (ours) gpus vs. $1$ (theirs).

\begin{table}[h]
    \centering \small 
    \begin{tabular}{p{3cm}llll}
\toprule
\bf Method & \multicolumn{2}{c}{\bf  ne-en} & \multicolumn{2}{c}{\bf  si-en} \\\cmidrule(r){2-3}\cmidrule(l){4-5}
& \bf 1M & \bf 5M & \bf 1M & \bf 5M\\
\midrule
        Main - Ensemble & 6.8 &  2.8 &{\bf  6.4} & 4.0 \\
        Constr. - \laser{} \emph{local} & {\bf 6.9} &  2.5 & 6.2 &3.8 \\ 
        Best (other) & 5.5 & {\bf 3.4} &5.0 & {\bf 4.4} \\
        \bottomrule
    \end{tabular}
    \caption{Official results of the main and secondary submissions on the \flores{} test set evaluated with the NMT configuration. For comparison, we include the best scores by another system.}
    \label{tab:official}
\end{table}

\subsection{Discussion}
One natural question to explore is how would the \laser{} method benefit if it had access to additional data. To explore this, we used the \laser{} open-source toolkit, which provides a trained encoder covering $93$ languages, but does not include Nepali. 
In Table~\ref{tab:pretrainlaser}, we observe that the pretrained \laser{} model outperforms the \laser{} \emph{local} model by $0.4$ BLEU. For \NeEn the situation reverses: \laser{} \emph{local} provides much better results. However, the results of the pretrained \laser{} are only slightly worse that those of Bicleaner (6.12) which is the best non-\laser{} method.  This suggests that \laser{} can function well in zero-shot scenarios (i.e. \NeEnns), but it works even better when it has additional supervision for the languages it is being tested on. 

\begin{table}[h]
    \centering \small 
    \begin{tabular}{p{3cm}llll}
\toprule
\bf Method & \multicolumn{2}{c}{\bf  ne-en} & \multicolumn{2}{c}{\bf  si-en} \\\cmidrule(r){2-3}\cmidrule(l){4-5}
& \bf 1M & \bf 5M & \bf 1M & \bf 5M\\
\midrule
        Pre-trained \laser{} & 6.06 & 1.49 &{\bf  7.82} & {\bf 5.56} \\
        \laser{} \emph{local} & {\bf 7.37} & {\bf 3.15} & 7.49 & 5.01 \\ 

        \bottomrule
    \end{tabular}
    \caption{Comparison of results on the \flores{} devtest set using the constrained and the pre-trained vesions of \laser{}.}
    \label{tab:pretrainlaser}
\end{table}

\section{Conclusions and Future Work}
In this paper, we describe our submission to the WMT low-resource parallel corpus filtering task. 
We use of multilingual sentence embeddings from \laser{} to filter noisy sentences. 
We observe that \laser{} can obtain better results than the baselines by a wide margin. Incorporating scores from other techniques and creating an ensemble provides additional gains. Our main submission to the shared task is based on the best of the ensemble configuration and our contrastive submission is based on the best \laser{} configuration. 
Our systems perform the best on the 1M condition for the \NeEn and \SiEn tasks.
We analyze the performance of a pre-trained version of \laser{} and observe that it can perform the filtering task well even in zero-resource scenarios, which is very promising.  

In the future, we want to evaluate this technique for high-resource scenarios and observe whether the same results transfer to that condition.
Moreover we plan to investigate how the size of training data (parallel, monolingual) impact low-resource sentence filtering task.

\bibliography{enne_ensi_wikidatasets}
\bibliographystyle{acl_natbib}
\end{document}